\pdfoutput=1  
\documentclass[runningheads]{llncs}
\usepackage[T1]{fontenc}
\usepackage{graphicx,verbatim}
\usepackage{xcolor} 
\usepackage{amsmath}
\usepackage{amssymb}
\usepackage{amsfonts}
\usepackage{booktabs}
\usepackage{xcolor,soul}
\usepackage[inline]{enumitem}

\usepackage{url}
\usepackage[capitalize]{cleveref}
\crefname{section}{Sec.}{Secs.}
\Crefname{section}{Section}{Sections}
\Crefname{table}{Table}{Tables}
\crefname{table}{Tab.}{Tabs.}

\begin{document}
%

\title{Cross-Modality Structural Guidance in 3D Latent Diffusion for Robust FLAIR Super-Resolution}

%
\author{Haoyu Lan\inst{1} \and
Jiazhen Zhang\inst{2} \and
John Onofrey\inst{2} \and
Bino Varghese\inst{1} \and
Nasim Sheikh-Bahaei\inst{1} \and
Arthur W. Toga\inst{1} \and
Jeiran Choupan\inst{1}}
\authorrunning{H. Lan et al.}
%
\institute{University of Southern California \\
\email{haoyulan@usc.edu, Bino.Varghese@med.usc.edu, Nasim.Sheikh-Bahaei@med.usc.edu, toga@loni.usc.edu, choupan@usc.edu} \and
Yale University \\
\email{\{jiazhen.zhang, john.onofrey\}@yale.edu}}

\titlerunning{Guided 3D Latent Diffusion for FLAIR Super-Resolution}

\maketitle              

\begingroup
\renewcommand\thefootnote{}%
\footnotetext{Accepted for oral presentation at MLCN 2026, a MICCAI 2026 workshop.
The Version of Record will appear in Springer Lecture Notes in Computer Science.}%
\endgroup

\begin{abstract}
High-resolution (HR) MRI acquisition is often hampered by scan time constraints, resulting in anisotropic or low-resolution scans (e.g., thick-slice FLAIR) that limit diagnostic accuracy. 
While deep learning–based super-resolution (SR) methods show promise, they often hallucinate anatomical details, which can compromise brain structural integrity. 
To mitigate this limitation, we introduce \textbf{MR-DiffuSR}, a \textbf{M}ulti-\textbf{R}esolution \textbf{Diffu}sion-based \textbf{S}uper-\textbf{R}esolution framework that incorporates HR T1w structural image priors to guide the restoration of thick-slice FLAIR scans and operates in the 3D latent space. 
Our architecture introduces cross-modality structural swin attention, which derives structural attention maps from the HR T1w and applies them to the low-resolution FLAIR latent features.
This design disentangles anatomical structure from modality-specific contrast, effectively preventing hallucinations. Furthermore, we employ a mixed-scale degradation strategy, training the model on a continuum of downsampling factors to ensure robustness to varying slice thicknesses, while optimizing with a DINOv3-based perceptual loss to preserve high-frequency semantic details. 
Evaluated on the ADNI-4 and ADNI-2 datasets, MR-DiffuSR surpasses both CNN and 2D diffusion approaches, achieving an average PSNR of 32.46\,dB, SSIM of 0.97, and LPIPS of 0.07 across all downsampling factors. In downstream white matter hyperintensity segmentation, our model demonstrates exceptional robustness. 
While baseline performance collapses at 10$\times$ downsampling (Dice: 0.51), MR-DiffuSR maintains a Dice score of 0.63, preserving utility even at 7\,mm equivalent slice thickness.





\end{abstract}

\begin{figure}[t]
    \centering
    \vspace{-1.0\baselineskip}
    \includegraphics[width=0.8\columnwidth]{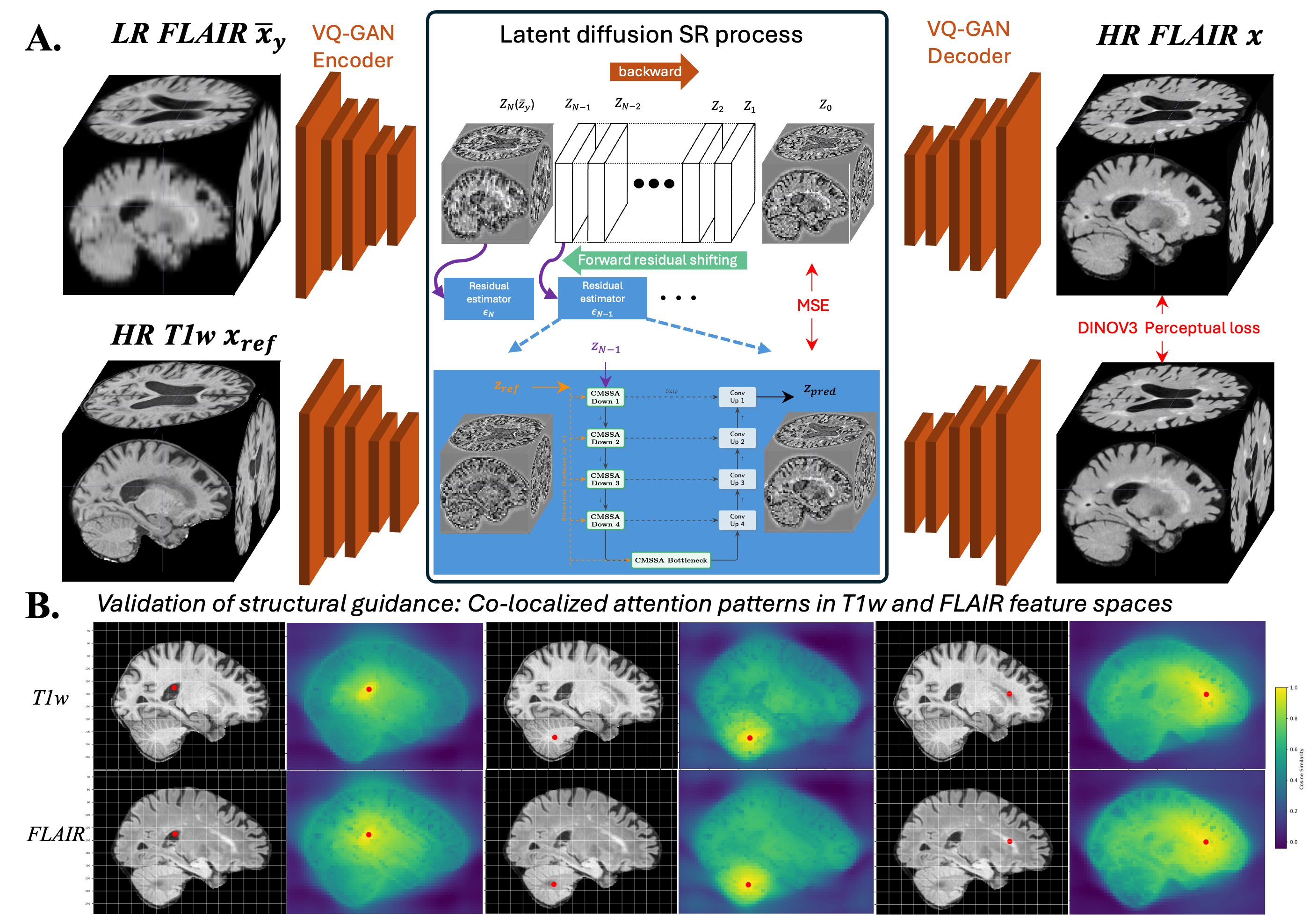}
    \caption{
    \textbf{Overview of MR-DiffuSR.} 
    (A) Architecture of the proposed framework, including a 3D VQ-GAN~\cite{esser2021taming} for latent compression, a T1w-guided attention module for structural conditioning, a residual-guided diffusion restoration model~\cite{yue2024efficient}, and a DINOv3–based perceptual regularizer~\cite{simeoni2025dinov3}. 
    (B) Visualization of consistent self-similarity patterns (with probing points) across T1w and FLAIR using DINOv3 output features, demonstrating anatomical congruence despite contrast differences.
    }
    \vspace{-1.5\baselineskip}
    \label{fig:figure1}
\end{figure}

\section{Introduction}

High-resolution (HR) MRI enables the detection of neurodegenerative pathology before regional atrophy occurs~\cite{frisoni2010clinical}. 
Specifically, HR T2 weighted fluid-attenuated inversion recovery (FLAIR) is critical for assessing white matter hyperintensities (WMH), a microscopic marker of small vessel disease, enabling early lesion identification and precise longitudinal tracking~\cite{wardlaw2013neuroimaging,debette2010clinical}. 
However, clinical routines often dictate anisotropic, thick-slice acquisitions (e.g., 5-7\,mm). This poor through-plane resolution impairs downstream lesion segmentation and prevents the integration of abundant legacy 2D thick-slice FLAIRs with 3D isotropic T1w for multi-modal analysis~\cite{greenspan2002mri}.

Super-resolution (SR) offers a retrospective solution to bridge the gap between clinical acquisition and high-fidelity analysis in medical imaging~\cite{van2012super,amoros2025evaluating,lu2024diffusion,giraldo2024perceptual,zhang2025deep}. 
Recent study~\cite{giraldo2024perceptual} shows that combining perceptual and reconstruction losses with physics-informed low-resolution (LR) simulation improves FLAIR SR.
While deep learning-based SR has made strong progress, existing methods face two critical limitations when applied to medical imaging. 
First, \textit{hallucination}: generative models often produce high-frequency textures that appear visually plausible but are anatomically incorrect~\cite{cohen2018distribution,giraldo2024perceptual,iglesias2023synthsr}.
This issue becomes more pronounced as the LR–HR gap increases, raising concerns about diagnostic reliability. 
Second, \textit{rigidity}: most SR models are trained on fixed degradation levels.
When applied to real-world data with varying slice thicknesses (e.g., legacy scans or multi-site cohorts)~\cite{iglesias2023synthsr}, their performance can degrade substantially.


To address these challenges, we propose \textbf{MR-DiffuSR}, a \textbf{M}ulti-\textbf{R}esolution \textbf{Diffu}sion-based \textbf{S}uper-\textbf{R}esolution framework that incorporates latent 3D compression, residual-guided diffusion restoration, and a DINOv3-based perceptual regularization term.
Compared with pure 2D SR methods, we ensure volumetric consistency using a latent 3D model with efficient computation. Unlike single-image SR approaches, our method leverages the routinely acquired HR T1-weighted (T1w) scan as a structural prior to guide reconstruction. We avoid direct T1w-to-FLAIR synthesis, which inherently risks hallucination or elimination of unpredictable pathology~\cite{cohen2018distribution}. 
Instead, MR-DiffuSR uses the HR T1w image strictly as a structural scaffold while solely relying on the LR FLAIR input to dictate true tissue contrast. We achieve this with a computationally efficient Cross-Modality Structural Swin Attention \textbf{(CMSSA)} mechanism, which leverages window-based attention to explicitly disentangle T1w structural geometry from FLAIR-specific contrast with low computational overhead. By anchoring the generative process to T1w anatomy and trained with scale-invariance (4$\times$--10$\times$), MR-DiffuSR achieves high-fidelity image restoration that remains robust under severe degradation. We evaluate MR-DiffuSR on the ADNI-4 dataset across multiple downsampling factors using standard image-quality metrics and downstream WMH segmentation~\cite{schmidt2017bayesian}.
MR-DiffuSR consistently improves reconstruction quality and yields higher segmentation accuracy than strong baselines, including SuperSynth~\cite{liu2025modality}, PRETTIER~\cite{giraldo2024perceptual}, and 2D diffusion-based SR variants, even at 7\,mm slice thickness.

\section{Methods}

    \subsection{Problem Formulation and MR-Physics Motivated Degradation}
    \label{sec:simu}
    We address the problem of multi-scale super-resolution, where the goal is to recover a high-resolution target volume $x$ from a low-resolution observation $y$ with resolution degradation. 
    Let $x \in \mathbb{R}^{H\times W\times D}$ denote an HR FLAIR volume and $x_{\mathrm{ref}} \in \mathbb{R}^{H\times W\times D}$ denote its co-registered HR T1w volume. 
    We observe a LR FLAIR volume $y$ generated by an acquisition-inspired degradation operator $\mathcal{A}_s(\cdot)$ parameterized by a downsampling factor $s$:
    \begin{equation}
        y = \mathcal{A}_s(x) + \epsilon, \quad \epsilon \sim \mathcal{N}(0,\sigma^2).
    \label{eq:forward}
    \end{equation}
    Given $(y,x_{\mathrm{ref}})$, our objective is to reconstruct $\hat{x} \approx x$. 
    To enable paired supervision, we simulate LR FLAIR volumes from HR data using an acquisition-inspired model~\cite{poot2010general}. 
    We model through-plane degradation as $\mathcal{A}_s(x)=\mathcal{D}_s\!\big(\mathcal{B}_{\text{slice}}(x)\big)$, where $\mathcal{B}_{\text{slice}}$ applies a 1D slice-selection profile blur along the slice axis, i.e., windowed-sinc slice excitation, and $\mathcal{D}_s$ performs strided downsampling to match the spacing for factor $s \in \{4, 6, 8, 10\}$. 
    This formulation approximates the partial volume effects (PVE) inherent in thick-slice acquisitions. 
    We assume access to a co-registered routine high-resolution T1w reference $x_{\mathrm{ref}}$, which provides complementary structural information with contrast distinct from the target FLAIR sequence.

    \subsection{Reference-Guided 3D Latent Diffusion Restoration}

    To efficiently model high-dimensional volumetric data and reduce computation, we operate in a compressed latent space. 
    We first train a 3D VQ-GAN~\cite{esser2021taming} using the HR FLAIR, then use its encoder $\mathcal{E}$ with quantization $Q(\mathcal{E}(\cdot))$ to map the input volumes into compact latent representations.
    The restoration process is formulated as a residual shifting (ResShift)~\cite{yue2024efficient} diffusion problem via a 3D Latent Diffusion Model (LDM)~\cite{rombach2022high}, with reference guidance.
    Following~\cite{yue2024efficient}, we construct a short Markov chain that transports a high-quality (HQ) latent toward its low-quality (LQ) counterpart by progressively \emph{shifting the residual}.

    \textbf{Latent initialization and Forward residual-shifting process.}
    Given an LR FLAIR observation $y$, we first upsample it to the HR grid ($\bar{x}_y = \mathrm{Up}(y)$) and encode it into a quantized latent ($\bar{z}_y = Q(\mathcal{E}(\bar{x}_y))$).
    Let $z_0 = Q(\mathcal{E}(x))$ denote the HQ latent of the HR target FLAIR volume $x$. We define the latent residual LQ$\rightarrow$HQ) as 
    $e_0 = \bar{z}_y - z_0.$
    
    We introduce a monotone shifting sequence $\{\eta_t\}_{t=1}^T$ with $\eta_1 \!\rightarrow\! 0$ and $\eta_T \!\rightarrow\! 1$, and define $\alpha_1=\eta_1$, $\alpha_t=\eta_t-\eta_{t-1}$ for $t>1$. The forward transition is
    \begin{equation}
    q(z_t \mid z_{t-1}, \bar{z}_y) = \mathcal{N}\!\big(z_t;\; z_{t-1} + \alpha_t e_0,\; \kappa^2 \alpha_t I\big), \quad t=1,\dots,T,
    \label{eq:forward_reshift}
    \end{equation}
    which yields the closed-form marginal
    \begin{equation}
    q(z_t \mid z_0, \bar{z}_y) = \mathcal{N}\!\big(z_t;\; z_0 + \eta_t e_0,\; \kappa^2 \eta_t I\big).
    \label{eq:marginal_reshift}
    \end{equation}
    
    \textbf{Reverse restoration with structural conditioning.}
    At inference, we start from a perturbed LQ-centered latent $z_T \sim \mathcal{N}(\bar{z}_y,\kappa^2 I)$ and iteratively sample along the reverse chain to recover $z_0$. We parameterize the reverse mean using a network $f_\theta$ that predicts the clean HQ latent:
    \begin{equation}
    p_\theta(z_{t-1}\mid z_t,\bar{z}_y,t) = \mathcal{N}\!\big(z_{t-1};\, \mu_\theta(z_t,\bar{z}_y,t),\, \kappa^2 \tfrac{\eta_{t-1}}{\eta_t}\alpha_t I\big),
    \label{eq:reverse_kernel}
    \end{equation}
    \begin{equation}
    \mu_\theta(z_t,\bar{z}_y,t)=\tfrac{\eta_{t-1}}{\eta_t} z_t + \tfrac{\alpha_t}{\eta_t} f_\theta(z_t,\bar{z}_y,t, z_{\mathrm{ref}}),
    \label{eq:reverse_mean}
    \end{equation}
    where $z_{\mathrm{ref}}=Q(\mathcal{E}(x_{\mathrm{ref}}))$ is the latent of HR T1w. In our implementation, $f_\theta$ is a 3D U-Net with CMSSA blocks (\cref{sec:cmssa}), so the reverse updates are anchored to T1w anatomy while preserving FLAIR contrast.
    
    \textbf{Reconstruction.}
    Residual-shifting formulation allows fast sampling with small $T$ while keeping the diffusion trajectory centered around the observed LR input.
    After sampling, we decode $\hat{x}=G(\hat{z}_0)$ with the VQ-GAN decoder $G(\cdot)$.
    
    \textbf{Cross modality structural swin attention (CMSSA).}
    \label{sec:cmssa}
    To restore missing high-frequency details, we introduce CMSSA (\cref{fig:figure1}A) to explicitly disentangle \textit{structural geometry} from \textit{modality-specific contrast}. 
    As evidenced by co-localized attention patterns in DINOv3 feature spaces (\cref{fig:figure1}B), T1w and FLAIR maintain strong anatomical congruence despite contrast differences. 
    Leveraging this, CMSSA uses the HR T1w latent feature ($F_{\mathrm{ref}} = Q(\mathcal{E}(x_{\mathrm{ref}}))$) as a structural scaffold to guide LR FLAIR latent feature ($F_{LR}=Q(\mathcal{E}(\mathrm{Up}(y))$) aggregation.
    
    To improve computational efficiency in attention calculation, the spatially aligned features are first partitioned into 3D windows. 
    Crucially, Queries ($\mathcal{Q}$) and Keys ($K$) are derived exclusively from $F_{Ref}$ to capture high-fidelity anatomical topology within the window partitions, while Values ($V$) are derived from $F_{LR}$ to preserve target contrast:
    \begin{equation}
        \text{CMSSA}(\mathcal{Q}, K, V) = \text{Softmax}\left(\frac{\mathcal{Q}K^T}{\sqrt{d_k}} + B\right)V,
    \end{equation}
    where $\mathcal{Q}=F_{Ref}W_Q$, $K=F_{Ref}W_K$, and $V=F_{LR}W_V$. $W_{\ast}$ are learnable projections, $d_k$ is the dimension of the attention head and $B$ is the relative position bias.
    This reference-driven attention, combined with standard alternating shifted-window (swin) partitioning~\cite{liu2021swin}, efficiently prevents hallucinations even under severe downsampling.

\begin{figure}[t!]
    \centering
    \includegraphics[width=0.9\columnwidth]{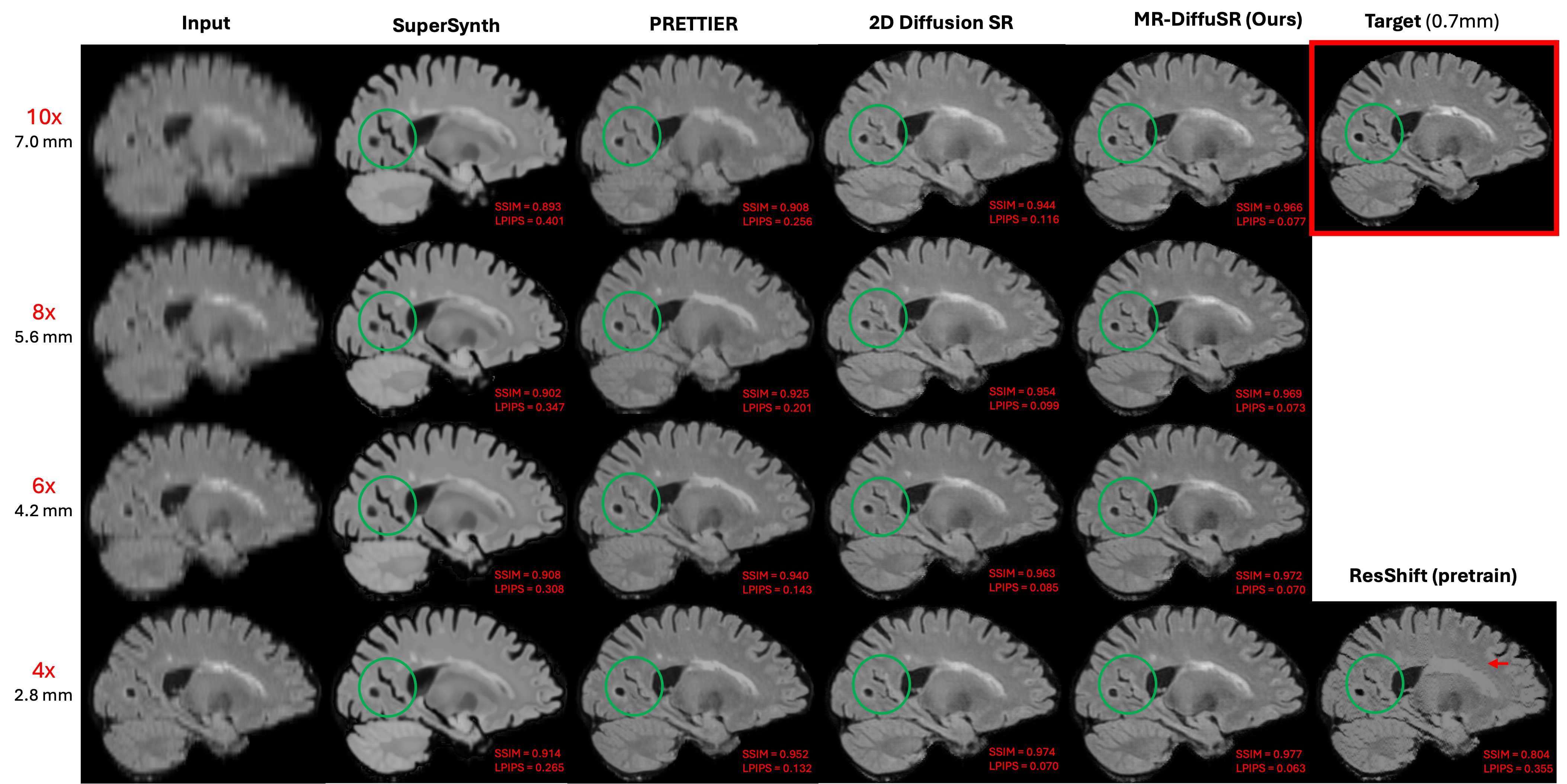}
    \vspace{-1.0\baselineskip}
    \caption{
    \textbf{Qualitative Results.} 
    Sagittal slices from a representative subject under different downsampling factors and reconstruction methods. 
    Highlighted regions (green circles) emphasize fine anatomical structures. 
    }
    \vspace{-1.5\baselineskip}
    \label{fig:figure2}
\end{figure}

    \subsection{Multi-Resolution Training and Perceptual Optimization}
    \label{dino}
    To improve robustness to varying slice thicknesses in clinical data, we adopt a multi-resolution training strategy. Rather than training at a fixed scale (e.g., $s=6$), we randomly sample through-plane degradation factors $s \in \{4,6,8,10\}$ during training. This encourages the model to learn a scale-consistent restoration mapping. At lower degradation levels (smaller $s$), the model relies more on the LR input itself, whereas at higher degradation levels (larger $s$) it increasingly leverages the T1w structural prior.

    For perceptual regularization, we utilize a pre-trained DINOv3~\cite{simeoni2025dinov3} encoder. Unlike VGG feature extractor~\cite{simonyan2014very} which focuses on texture, DINOv3 features capture dense semantic anatomical properties, ensuring that the restored structures are not just texturally sharp but semantically meaningful~\cite{liu2025does,ma2026pixelgen}.
    The perceptual loss is defined as
    $\mathcal{L}_{\text{dino}} =
    \sum_{l\in\mathcal{S}}
    \left\|\Phi_l(\hat{x})-\Phi_l(x)\right\|_1,
    \label{eq:dino_loss}$
    where $\Phi_l(\cdot)$ extracts DINOv3 features at selected layers $l$.
    The network is optimized using a hybrid loss function 
    $\mathcal{L} = \mathcal{L}_{\text{diff}} + \lambda_{\text{dino}}\mathcal{L}_{\text{dino}},
    \label{eq:total_loss}$
    where $\mathcal{L}_{\text{diff}}$ is the mean square error (MSE) loss of the latent diffusion process.


\begin{figure}[t!]
    \centering
    \includegraphics[width=1\columnwidth]{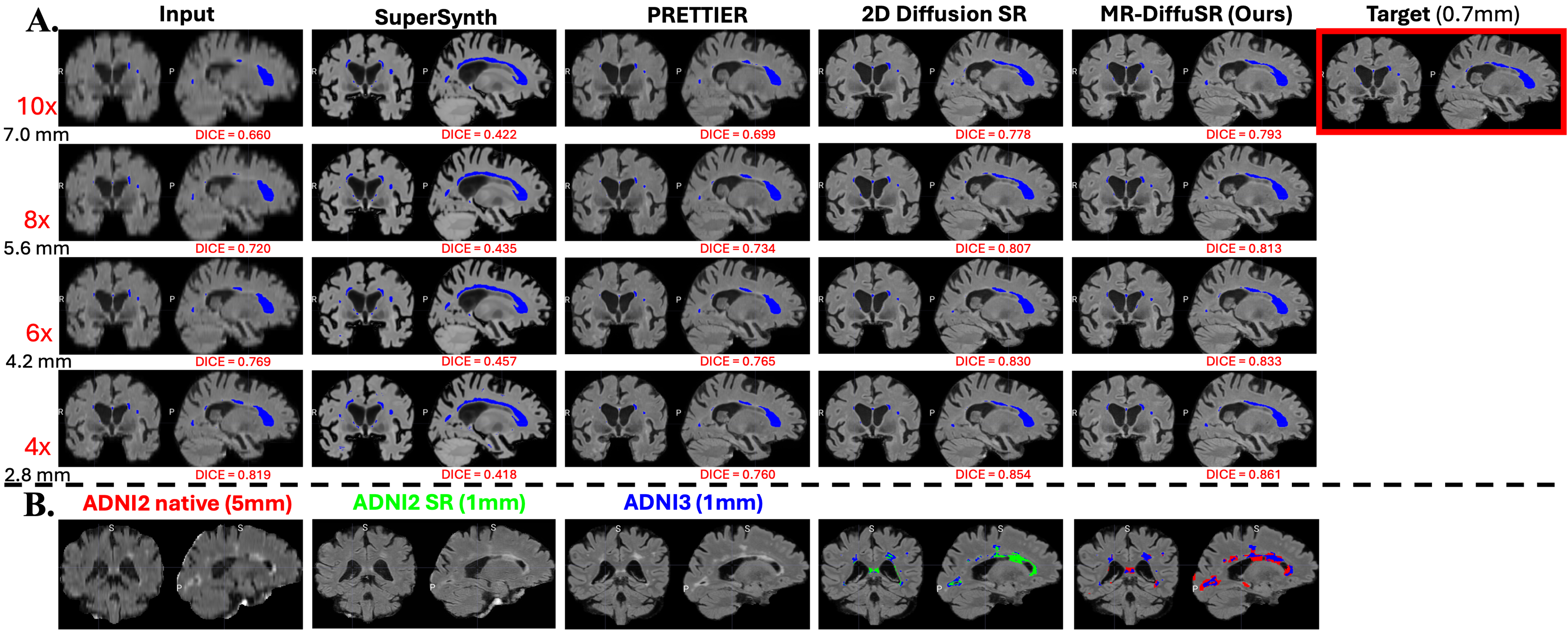}
    \vspace{-2.0\baselineskip}
    \caption{
    \textbf{WMH Segmentation.} 
    A: Multiscale masks (blue) demonstrate MR-DiffuSR preserves structure across methods and downsampling factors. B: Longitudinal evaluation. Color-coded masks (\textcolor{red}{Red: Native ADNI-2}; \textcolor{green}{Green: ADNI-2 SR}; \textcolor{blue}{Blue: ADNI-3 Follow-up}) show correction of PVE-induced overestimation.
    }
    \vspace{-2.0\baselineskip}
    \label{fig:figure3}
\end{figure}

\section{Experiments and Results}
    \subsection{Experimental Setup}
    \textbf{Dataset:} 
    We utilized the Alzheimer's Disease Neuroimaging Initiative 4 (ADNI-4) dataset~\cite{miller2024adni4}, consisting of 318 quality-controlled paired T1w and FLAIR volumes at 0.7\,mm isotropic resolution. 
    We used 249 subjects for training and held out 69 subjects for evaluation.
    Both T1w and FLAIR images were preprocessed with co-registration, skull stripping, intensity scaling, and bias correction with the N4 algorithm~\cite{tustison2010n4itk}. 
    LR inputs were simulated using a physics-informed degradation model (\cref{sec:simu}).
    For downstream clinical evaluation, we assess WMH segmentation using the lesion prediction algorithm (LPA)~\cite{schmidt2017bayesian}. 
    We treat the WMH mask obtained from the HR FLAIR as the gold standard and compare it against the masks produced from restored FLAIR outputs. 

    \textbf{Baselines:} 
    We compared MR-DiffuSR against three methods: 
    (1) \textbf{SuperSynth}~\cite{liu2025modality}, a real/synthetic data trained CT/MR foundation model; 
    (2) \textbf{PRETTIER}~\cite{giraldo2024perceptual}, a perceptual-optimized restoration framework; 
    and (3) \textbf{2D diffusion SR}, a standard 2D slice-wise diffusion model without structural guidance. 
    Additionally, we included a simple linear interpolation SR and a standard pre-trained restoration baseline (ResShift).
    
    \textbf{Experimental Settings:} 
    We first train a custom 3D VQ-GAN on the ADNI-4 dataset to encode the $256^3 \times 1$ input volumes into a $64^3 \times 32$ latent space (spatial $\times$ channel).
    We then train the diffusion model in the latent space for 100{,}000 iterations using AdamW~\cite{loshchilov2017decoupled} optimizer with a learning rate of $1\times 10^{-5}$, no weight decay, and $\lambda_{\text{dino}}=1$. 
    At inference, we use a 4-step exponential noise schedule ($\kappa=2.0$, $\sigma_{min}=0.01$). 
    Training was stabilized using automatic mixed precision (AMP) and an exponential moving average (EMA) rate of 0.999. 
    All experiments were implemented in Python 3.10.19 and PyTorch 2.2.2.
    
    \textbf{Evaluation Metrics:} 
    We report image quality using Normalized Root Mean Square Error (NRMSE $\downarrow$), Peak Signal-to-Noise Ratio (PSNR $\uparrow$), Structural Similarity Index (SSIM $\uparrow$), and Learned Perceptual Image Patch Similarity~\cite{zhang2018unreasonable} (LPIPS $\downarrow$) utilizing a DINOv3 backbone. 
    For WMH segmentation, we report Dice and Sensitivity.
    Wilcoxon signed-rank tests with Bonferroni correction evaluated significance of our method against each baseline model across all evaluation metrics, with $p$ < 0.05 considered significant.

\begin{figure}[t!]
    \centering
    \includegraphics[width=1.0\columnwidth]{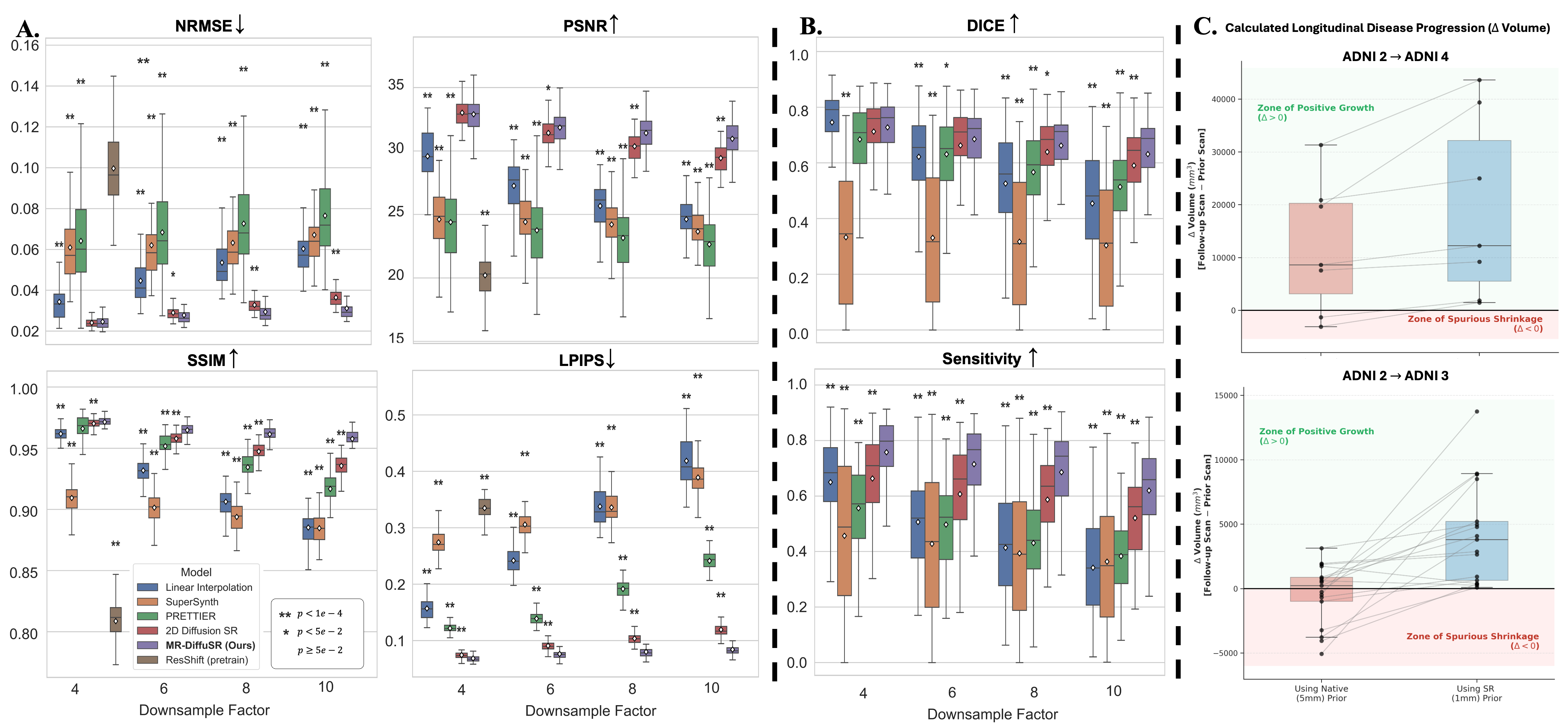}
    \vspace{-2.0\baselineskip}
    \caption{
    \textbf{Quantitative Results.} Box plots of (A) reconstruction and (B) WMH segmentation metrics across downsampling factors. Diamonds denote means ($^*p<0.05$, $^{**}p<0.0001$). (C) Calculated longitudinal WMH progression ($\Delta$ Volume) from legacy ADNI-2 (5mm) to ADNI-3/4 (1mm). Native 5mm baselines suffer from PVE-induced overestimation, causing biologically implausible ``spurious shrinkage'' ($\Delta V < 0$, red zone). MR-DiffuSR restores accurate baseline volumes, capturing reliable disease progression (green zone).
    }
    

    \vspace{-1.5\baselineskip}
    \label{fig:figure4}
\end{figure}
    
    \subsection{Results and Analysis}

    \textbf{Qualitative Results:}
    \cref{fig:figure2} shows representative reconstructions across downsampling factors. 
    At larger degradation factors (8$\times$–10$\times$), baseline methods over-smooth fine structures or produce inconsistent textures, especially around cortical boundaries and periventricular regions (highlighted). 
    In contrast, MR-DiffuSR better preserves structural continuity while maintaining FLAIR-specific contrast, yielding outputs that more closely approximate the 0.7\,mm target. 
    The red arrow in \cref{fig:figure2} highlights slice inconsistencies in the 2D method, which are resolved by our 3D VQ-GAN decoder.
    \cref{fig:figure3}A compares WMH segmentation overlays across downsampling factors. 
    As degradation increases, baseline SR outputs yield more fragmented or shifted WMH masks and lower Dice, especially at 8$\times$–10$\times$. 
    MR-DiffuSR better preserves lesion extent and spatial consistency, producing overlays closer to the 0.7\,mm target under severe downsampling.
    Furthermore, longitudinal evaluation (\cref{fig:figure3}B) color-coded masks (Native ADNI-2: Red; ADNI-2 SR: Green; ADNI-3 Follow-up: Blue) visually showcase that MR-DiffuSR effectively resolves PVE-induced overestimation on native legacy scans, preserving true disease progression.

    \begin{table}[htbp]
    \vspace{-4mm} %
    \caption{\textbf{Ablation study}. Impact of perceptual loss and CMSSA ($8\times$).}
    \label{tab:ablation}
    \vspace{-2mm} %
    \centering
    \footnotesize %
    \renewcommand{\arraystretch}{0.9} %
    \setlength{\tabcolsep}{4pt} %
    \begin{tabular}{l c c c c}
    \hline
    Model & NRMSE (\%) $\downarrow$ & PSNR $\uparrow$ & SSIM $\uparrow$ & LPIPS (\%) $\downarrow$ \\
    \hline
    Baseline (FLAIR only) & 3.45$\pm$0.42 & 29.92$\pm$1.20 & 0.94$\pm$0.01 & 10.45$\pm$1.07 \\
    + Perceptual Loss     & 3.43$\pm$0.44 & 29.98$\pm$1.19 & 0.94$\pm$0.01 & 10.38$\pm$1.10 \\
    + CMSSA (Ours)        & \textbf{2.80$\pm$0.36} & \textbf{31.74$\pm$1.20} & \textbf{0.96$\pm$0.01} & \textbf{7.84$\pm$0.89} \\
    \hline
    \end{tabular}
    \vspace{-5mm} %
\end{table}

    \textbf{Quantitative Results:} 
    \cref{fig:figure4}A shows consistent gains for MR-DiffuSR across all downsampling factors. 
    Notably, at the most challenging 10$\times$ downsampling factor ($\approx$ 7\,mm slice thickness), our method achieves a PSNR of \textbf{30.96\,dB} and LPIPS of \textbf{0.08}, compared to 22.6\,dB/0.24 for PRETTIER. 
    The box plots in \cref{fig:figure4}A show a narrower interquartile range for our method, indicating more consistent performance across subjects compared to the baselines.
    As downsampling increases, baseline performance degrades more sharply, while MR-DiffuSR remains stable, supporting the mixed-scale training strategy and the use of the T1w reference for structural anchoring (detailed $p$-values are provided in \cref{fig:figure4}).

    \cref{fig:figure4}B evaluates downstream clinical impact on WMH segmentation. 
    Segmentation accuracy on baseline reconstructions drops substantially at higher degradation (e.g., PRETTIER Dice 0.51 at 10$\times$), whereas MR-DiffuSR maintains 0.63 Dice, indicating better preservation of clinically specific contrast features required for automated pathology detection.

    Furthermore, to assess clinical utility, we evaluated longitudinal WMH progression using legacy ADNI-2 (5mm) scans with ADNI-3/4 (1mm) follow-ups (\cref{fig:figure4}C). Due to severe PVE, native 5mm baselines systematically overestimate initial WMH, producing biologically implausible "spurious shrinkage" ($\Delta V < 0$) over time. MR-DiffuSR effectively mitigates this PVE-induced overestimation, restoring biologically plausible longitudinal progression (positive growth) that aligns with true disease evolution. This confirms its value in super-resolving legacy cohorts for reliable longitudinal analysis.

    
    \textbf{Ablation Study:} 
    Directly reconstructing the LR features without diffusion yields only 26.32\,dB PSNR and 0.90 SSIM, highlighting the need for the diffusion stage.
    \cref{tab:ablation} summarizes a component-wise ablation.
    Starting from a 3D baseline trained with MSE only, adding perceptual loss leads to small but consistent improvements across image metrics.
    In contrast, introducing CMSSA guidance gives the largest gain, increasing PSNR from 29.92\,dB to 31.74\,dB and reducing LPIPS to 0.08. 
    Overall, perceptual loss mainly helps texture, while the T1w structural prior drives most of the reconstruction improvement.

\section{Discussion and Conclusion}




Our current results suggest three key takeaways: 
(1) T1 guidance provides a strong anatomical prior that becomes more critical at higher downsampling factors, mitigating the SR quality degradation typically observed where the LR–HR gap is larger.
(2) Latent 3D modeling improves volumetric consistency and computational efficiency relative to purely 2D models. 
(3) Improvements in image quality translate to better downstream lesion segmentation, supporting the potential for reliable quantitative analysis from routine clinical acquisitions.

Limitations include dependence on simulated degradations and the need to validate across broader acquisition protocols and multi-site clinical variability (slice gaps, sequences). Future work will extend evaluation to additional datasets and WMH segmentation pipelines, and explore efficiency optimizations for deployment. 
In conclusion, MR-DiffuSR offers a robust framework for retrospective FLAIR super-resolution, combining generative power with structural rigor to enable high-resolution analysis of thick-slice datasets.




\begin{credits}
\subsubsection{\ackname} This work was supported by the National Institutes of Health under grant numbers R01NS128486 and R01AG070825.

\subsubsection{\discintname}
The authors have no competing interests to declare that are relevant to the content of this article.
\end{credits}

%
%
%
\bibliographystyle{splncs04}
\bibliography{mybibliography}

@phdthesis{schmidt2017bayesian,
  title={Bayesian inference for structured additive regression models for large-scale problems with applications to medical imaging},
  author={Schmidt, Paul},
  year={2017},
  school={lmu}
}

@inproceedings{esser2021taming,
  title={Taming transformers for high-resolution image synthesis},
  author={Esser, Patrick and Rombach, Robin and Ommer, Bjorn},
  booktitle={Proceedings of the IEEE/CVF conference on computer vision and pattern recognition},
  pages={12873--12883},
  year={2021}
}

@article{yue2024efficient,
  title={Efficient diffusion model for image restoration by residual shifting},
  author={Yue, Zongsheng and Wang, Jianyi and Loy, Chen Change},
  journal={IEEE Transactions on Pattern Analysis and Machine Intelligence},
  year={2024},
  publisher={IEEE}
}

@article{simeoni2025dinov3,
  title={Dinov3},
  author={Sim{\'e}oni, Oriane and Vo, Huy V and Seitzer, Maximilian and Baldassarre, Federico and Oquab, Maxime and Jose, Cijo and Khalidov, Vasil and Szafraniec, Marc and Yi, Seungeun and Ramamonjisoa, Micha{\"e}l and others},
  journal={arXiv preprint arXiv:2508.10104},
  year={2025}
}

@article{liu2025modality,
  title={A modality-agnostic multi-task foundation model for human brain imaging},
  author={Liu, Peirong and Puonti, Oula and Hu, Xiaoling and Gopinath, Karthik and Sorby-Adams, Annabel and Alexander, Daniel C and Kimberly, W Taylor and Iglesias, Juan E},
  journal={arXiv preprint arXiv:2509.00549},
  year={2025}
}

@article{giraldo2024perceptual,
  title={Perceptual super-resolution in multiple sclerosis MRI},
  author={Giraldo, Diana L and Khan, Hamza and Pineda, Gustavo and Liang, Zhihua and Lozano-Castillo, Alfonso and Van Wijmeersch, Bart and Woodruff, Henry C and Lambin, Philippe and Romero, Eduardo and Peeters, Liesbet M and others},
  journal={Frontiers in Neuroscience},
  volume={18},
  pages={1473132},
  year={2024},
  publisher={Frontiers Media SA}
}

@inproceedings{poot2010general,
  title={General and efficient super-resolution method for multi-slice MRI},
  author={Poot, Dirk HJ and Van Meir, Vincent and Sijbers, Jan},
  booktitle={International Conference on Medical Image Computing and Computer-Assisted Intervention},
  pages={615--622},
  year={2010},
  organization={Springer}
}

@inproceedings{rombach2022high,
  title={High-resolution image synthesis with latent diffusion models},
  author={Rombach, Robin and Blattmann, Andreas and Lorenz, Dominik and Esser, Patrick and Ommer, Bj{\"o}rn},
  booktitle={Proceedings of the IEEE/CVF conference on computer vision and pattern recognition},
  pages={10684--10695},
  year={2022}
}

@inproceedings{liu2021swin,
  title={Swin transformer: Hierarchical vision transformer using shifted windows},
  author={Liu, Ze and Lin, Yutong and Cao, Yue and Hu, Han and Wei, Yixuan and Zhang, Zheng and Lin, Stephen and Guo, Baining},
  booktitle={Proceedings of the IEEE/CVF international conference on computer vision},
  pages={10012--10022},
  year={2021}
}

@article{liu2025does,
  title={Does DINOv3 set a new medical vision standard?},
  author={Liu, Che and Chen, Yinda and Shi, Haoyuan and Lu, Jinpeng and Jian, Bailiang and Pan, Jiazhen and Cai, Linghan and Wang, Jiayi and Zhang, Yundi and Li, Jun and others},
  journal={arXiv e-prints},
  pages={arXiv--2509},
  year={2025}
}

@article{ma2026pixelgen,
  title={PixelGen: Pixel Diffusion Beats Latent Diffusion with Perceptual Loss},
  author={Ma, Zehong and Xu, Ruihan and Zhang, Shiliang},
  journal={arXiv preprint arXiv:2602.02493},
  year={2026}
}

@article{lu2024diffusion,
  title={Diffusion-based deep learning method for augmenting ultrastructural imaging and volume electron microscopy},
  author={Lu, Chixiang and Chen, Kai and Qiu, Heng and Chen, Xiaojun and Chen, Gu and Qi, Xiaojuan and Jiang, Haibo},
  journal={Nature communications},
  volume={15},
  number={1},
  pages={4677},
  year={2024},
  publisher={Nature Publishing Group UK London}
}

@article{amoros2025evaluating,
  title={Evaluating super-resolution models in biomedical imaging: applications and performance in segmentation and classification},
  author={Amoros, Mario and Curado, Manuel and Vicent, Jose F},
  journal={Journal of Imaging},
  volume={11},
  number={4},
  pages={104},
  year={2025},
  publisher={MDPI}
}

@article{zhang2025deep,
  title={Deep learning--based super-resolution reconstruction on undersampled brain diffusion-weighted MRI for infarction stroke: a comparison to conventional iterative reconstruction},
  author={Zhang, Shuo and Zhong, Meimeng and Shenliu, Hanxu and Wang, Nan and Hu, Shuai and Lu, Xulun and Lin, Liangjie and Zhang, Haonan and Zhao, Yan and Yang, Chao and others},
  journal={American Journal of Neuroradiology},
  volume={46},
  number={1},
  pages={41--48},
  year={2025},
  publisher={American Journal of Neuroradiology}
}

@article{van2012super,
  title={Super-resolution in magnetic resonance imaging: a review},
  author={Van Reeth, Eric and Tham, Ivan WK and Tan, Cher Heng and Poh, Chueh Loo},
  journal={Concepts in Magnetic Resonance Part A},
  volume={40},
  number={6},
  pages={306--325},
  year={2012},
  publisher={Wiley Online Library}
}

@article{tustison2010n4itk,
  title={N4ITK: improved N3 bias correction},
  author={Tustison, Nicholas J and Avants, Brian B and Cook, Philip A and Zheng, Yuanjie and Egan, Alexander and Yushkevich, Paul A and Gee, James C},
  journal={IEEE transactions on medical imaging},
  volume={29},
  number={6},
  pages={1310--1320},
  year={2010},
  publisher={IEEE}
}

@article{miller2024adni4,
  title={The ADNI4 Digital Study: A novel approach to recruitment, screening, and assessment of participants for AD clinical research},
  author={Miller, Melanie J and Diaz, Adam and Conti, Catherine and Albala, Bruce and Flenniken, Derek and Fockler, Juliet and Kwang, Winnie and Sacrey, Diana Truran and Ashford, Miriam T and Skirrow, Caroline and others},
  journal={Alzheimer's \& Dementia},
  volume={20},
  number={10},
  pages={7232--7247},
  year={2024},
  publisher={Wiley Online Library}
}

@article{wardlaw2013neuroimaging,
  title={Neuroimaging standards for research into small vessel disease and its contribution to ageing and neurodegeneration},
  author={Wardlaw, Joanna M and Smith, Eric E and Biessels, Geert J and Cordonnier, Charlotte and Fazekas, Franz and Frayne, Richard and Lindley, Richard I and T O'Brien, John and Barkhof, Frederik and Benavente, Oscar R and others},
  journal={The Lancet Neurology},
  volume={12},
  number={8},
  pages={822--838},
  year={2013},
  publisher={Elsevier}
}

@article{frisoni2010clinical,
  title={The clinical use of structural MRI in Alzheimer disease},
  author={Frisoni, Giovanni B and Fox, Nick C and Jack Jr, Clifford R and Scheltens, Philip and Thompson, Paul M},
  journal={Nature reviews neurology},
  volume={6},
  number={2},
  pages={67--77},
  year={2010},
  publisher={Nature Publishing Group UK London}
}

@article{debette2010clinical,
  title={The clinical importance of white matter hyperintensities on brain magnetic resonance imaging: systematic review and meta-analysis},
  author={Debette, St{\'e}phanie and Markus, HS20660506},
  journal={Bmj},
  volume={341},
  year={2010},
  publisher={British Medical Journal Publishing Group}
}

@article{greenspan2002mri,
  title={MRI inter-slice reconstruction using super-resolution},
  author={Greenspan, Hayit and Oz, G and Kiryati, N and Peled, SLBG},
  journal={Magnetic resonance imaging},
  volume={20},
  number={5},
  pages={437--446},
  year={2002},
  publisher={Elsevier}
}

@article{simonyan2014very,
  title={Very deep convolutional networks for large-scale image recognition},
  author={Simonyan, Karen and Zisserman, Andrew},
  journal={arXiv preprint arXiv:1409.1556},
  year={2014}
}

@article{loshchilov2017decoupled,
  title={Decoupled weight decay regularization},
  author={Loshchilov, Ilya and Hutter, Frank},
  journal={arXiv preprint arXiv:1711.05101},
  year={2017}
}

@inproceedings{zhang2018unreasonable,
  title={The unreasonable effectiveness of deep features as a perceptual metric},
  author={Zhang, Richard and Isola, Phillip and Efros, Alexei A and Shechtman, Eli and Wang, Oliver},
  booktitle={Proceedings of the IEEE conference on computer vision and pattern recognition},
  pages={586--595},
  year={2018}
}

@article{iglesias2023synthsr,
  title={SynthSR: A public AI tool to turn heterogeneous clinical brain scans into high-resolution T1-weighted images for 3D morphometry},
  author={Iglesias, Juan E and Billot, Benjamin and Balbastre, Ya{\"e}l and Magdamo, Colin and Arnold, Steven E and Das, Sudeshna and Edlow, Brian L and Alexander, Daniel C and Golland, Polina and Fischl, Bruce},
  journal={Science advances},
  volume={9},
  number={5},
  pages={eadd3607},
  year={2023},
  publisher={American Association for the Advancement of Science}
}

@inproceedings{cohen2018distribution,
  title={Distribution matching losses can hallucinate features in medical image translation},
  author={Cohen, Joseph Paul and Luck, Margaux and Honari, Sina},
  booktitle={International conference on medical image computing and computer-assisted intervention},
  pages={529--536},
  year={2018},
  organization={Springer}
}
%





\end{document}